\documentclass[lettersize,journal]{IEEEtran}
\usepackage{amsmath,amsfonts}
\usepackage{algorithmic}
\usepackage{algorithm}
\usepackage{array}
\usepackage[caption=false,font=normalsize,labelfont=sf,textfont=sf]{subfig}
\usepackage{textcomp}
\usepackage{stfloats}
\usepackage{url}
\usepackage{verbatim}
\usepackage{graphicx}
\usepackage{cite}
\usepackage{bbding}
\usepackage{color}
\begin{document}

\title{Predicting Scores of Various Aesthetic Attribute Sets by Learning from Overall Score Labels}

\author{\IEEEauthorblockN{
		Heng Huang\IEEEauthorrefmark{1}\IEEEauthorrefmark{2},
		Xin Jin\IEEEauthorrefmark{2},
		Yaqi Liu\IEEEauthorrefmark{2}, 
		Hao Lou\IEEEauthorrefmark{2}, 
		Chaoen Xiao\IEEEauthorrefmark{2}, 
		Shuai Cui\IEEEauthorrefmark{3}, 
		Xinning Li\IEEEauthorrefmark{2}, 
		Dongqing Zou\IEEEauthorrefmark{4}}
	\IEEEauthorblockA{\IEEEauthorrefmark{1}University of Science and Technology of China, Anhui {\rm 230026}, China\\}
    \IEEEauthorblockA{\IEEEauthorrefmark{2}Beijing Electronic Science and Technology Institute, Beijing {\rm 100070}, China\\}
    \IEEEauthorblockA{\IEEEauthorrefmark{3}University of Edinburgh, Edinburgh {\rm EH89YL}, The U.K.\\}
    \IEEEauthorblockA{\IEEEauthorrefmark{4}Sensetime Research, Beijing {\rm 100089}, China}
\thanks{Heng Huang and Xin Jin are corresponding authors (email: hecate@mail.ustc.edu.cn, jinxinbesti@foxmail.com)}
}


\markboth{Journal of \LaTeX\ Class Files,~Vol.~14, No.~8, August~2021}%
{Heng Huang, Xin Jin \MakeLowercase{\textit{et al.}}: Predicting Scores of Various Aesthetic Attribute Sets by Learning from Overall Score Labels}


\maketitle

\begin{abstract}
Now many mobile phones embed deep-learning models for evaluation or guidance on photography. These models cannot provide detailed results like human pose scores or scene color scores because of the rare of corresponding aesthetic attribute data. However, the annotation of image aesthetic attribute scores requires experienced artists and professional photographers, which hinders the collection of large-scale fully-annotated datasets. In this paper, we propose to replace image attribute labels with feature extractors. First, a novel aesthetic attribute evaluation framework based on attribute features is proposed to predict attribute scores and overall scores. We call it the F2S (attribute features to attribute scores) model. We use networks from different tasks to provide attribute features to our F2S models. Then, we define an aesthetic attribute contribution to describe the role of aesthetic attributes throughout an image and use it with the attribute scores and the overall scores to train our F2S model. Sufficient experiments on publicly available datasets demonstrate that our F2S model achieves comparable performance with those trained on the datasets with fully-annotated aesthetic attribute score labels. Our method makes it feasible to learn meaningful attribute scores for various aesthetic attribute sets in different types of images with only overall aesthetic scores. 
\end{abstract}


\section{Introduction}
In image attribute aesthetics, image aesthetic datasets are usually explored for attribute evaluation. There are several frequently used datasets, e.g., AVA~\cite{murray2012ava}, AADB~\cite{kong2016photo}, EVA~\cite{kang2020eva}. However, those datasets cannot meet all our needs for different reasons. For example, there is no attribute evaluation in AVA; the number of the commenter in AADB is too small; EVA only has a few basic attributes. If researchers attempt to study image aesthetic attributes more deeply or investigate aesthetic attributes from different aspects, they even cannot find a publicly available dataset.

Some researchers hire professional workers or recruit volunteers to score attributes. Although they put a great deal of effort into attribute scoring, the results are not good enough. First, the evaluators need to have enough knowledge about aesthetics. Second, each image needs to be scored by different evaluators to reduce the influences of personality. Last but not least, intensive work may decrease the judgment of evaluators, and they need enough time to relax. In summary, manual attribute scoring seems to be unworkable for large-scale attribute scoring. 

\begin{figure}[htbp]
	\centering
	\includegraphics[width=\linewidth]{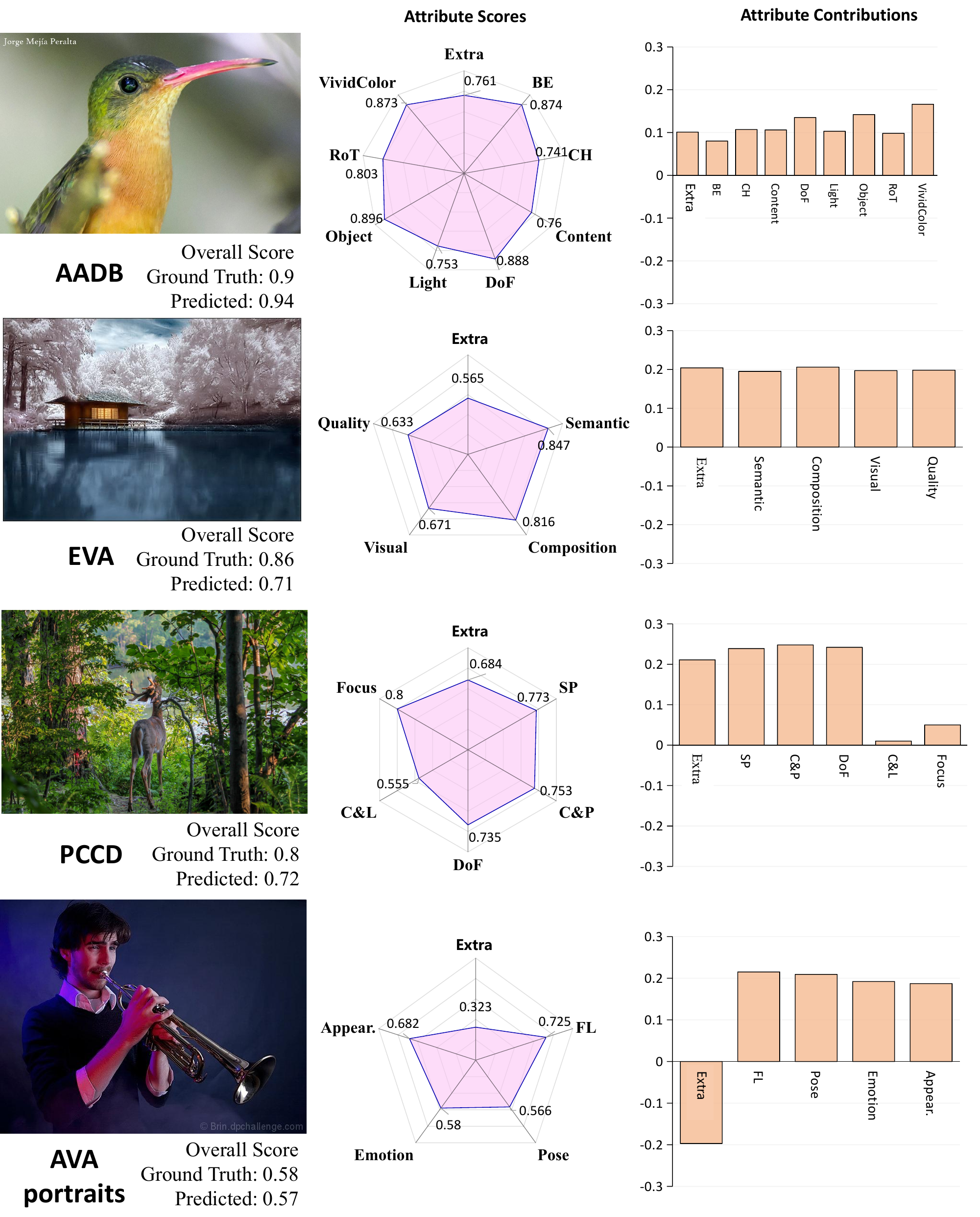}
	\caption{Visual examples of attribute contributions and attribute scores predicted by our model. Besides attribute scores, we also evaluate the contribution of each aesthetic attribute. The four examples come from different F2S models trained on AADB~\cite{kong2016photo}, EVA~\cite{kang2020eva}, PCCD~\cite{chang2017aesthetic}, and AVA~\cite{murray2012ava}. The attribute names in first three images come from the datasets. ``FL'' means ``face lighting''. ``Appear.'' means ``appearance''. In the first picture, ``object" and ``vivid color" has more contributions to the overall score. In the third picture, it shows that the ``C\&L" can be better to make the picture more beautiful. }
	\label{fig:oe2}
\end{figure}

In this paper, we propose a novel aesthetic attribute evaluation framework that trains the model to predict attribute scores without attribute labels. We borrow pre-trained image features from different tasks for aesthetic attribute extraction and scoring. For example, we adopt an object detection network to extract the object-level features and their spatial location information. We use these features as parts of attribute scoring input, and then these features are jointly trained with other carefully selected image features for attribute scoring and extra aesthetic scoring. With the help of object-level features, we can let our network know whether the objects and their corresponding locations are beautiful. We use a full-connection layer as the contribution module. This module establishes the relationship among different attribute features. Our model predicts the attribute scores and extra aesthetic scores but also outputs the attribute contributions by our attribute understanding attention module, as shown in Figure~\ref{fig:oe2}. Thus, we can get a more comprehensive aesthetic analysis for investigated images. 

In summary, our contribution can be concluded as follows:
\begin{itemize}
	\item A novel model of aesthetic attribute evaluation of images, called the 'F2S' model, is proposed to predict both the overall aesthetic and attribute scores. Our model is trained from images with only overall aesthetic score labels, while can predict attribute scores.
	\item Many models from other tasks are used to provide image features and we corroborate that these features reflect the aesthetic tendency.  
	\item Define aesthetic attribute contributions, and use an attention module to output contribution values. We reveal a method that generates attribute contribution values and attribute scores by learning from overall scores. The method includes a loss function to train the attribute scores from the overall score labels. The loss function represents the relationship among the attribute scores, the attribute contributions, and the overall score. The attributes can be adjusted in different types of images, such as portraits, buildings, and animals.
\end{itemize}

\section{Related Works}

Recently, several assistance systems~\cite{kim2019picme,rawat2016clicksmart,samsum,su2021camera} for mobile photography are proposed. Rawat et al.~\cite{rawat2016clicksmart} use photos from social media to suggest the best shooting place and angle for mobile photography. The shot suggestion function of the Samsung Galaxy S10~\cite{samsum} shows the guidance line and circle to guide users to obtain a good composition. Google proposes camera view adjustment prediction~\cite{su2021camera} for improving photo composition for mobile photography. The PicMe system proposed by Kim et al.~\cite{kim2019picme} uses interactive visual guidance for taking requested photo compositions by another user. The models rely on lots of image aesthetic data.

In the past decades, image aesthetics assessment was mainly through the human aesthetic perception of image features and photography rules. The features included the spatial distribution of edges, color distribution, hue and blur, etc.~\cite{luo2011content}. While drawing on some specific rules in photography, such as low depth-of-field indicator, colorfulness measure, shape convexity score and familiarity measure~\cite{DattaECCV2006}, Rule of Thirds etc.~\cite{MurrayCVPR2012}, with the development of feature extraction technology, features based on high-level aesthetic principles had emerged, such as features based on scene types and related contents~\cite{dhar2011high}, high-level semantic features based on this subject and background division~\cite{luo2008photo}.

With the continuous advancement of deep learning research in recent years, CNN-based deep learning models were widely used in the classification and regression of aesthetics. Kao et al.~\cite{kao2017deep} proposed the multi-task learning. They led the relevant items between tasks to the framework and made the utility of the appreciation of aesthetic and semantic labels more effective. Kong et al.~\cite{KongECCV2016} utilized the relative aesthetics to select the new datasets of image from the datasets with the pairs of related labels, and trained the related image pairs to obtain the aesthetic ranking with the higher accuracy. Sheng et al.~\cite{sheng2018attention} adopted attention-based multi-patch aggregation to adjust the weight of each patch in the training process. The results improved learning effectively. Pan et al.~\cite{pan2019image} proposed a multi-task depth convolution scoring network to learn aesthetic scores and attributes at the same time. The multi-task depth network wants to output aesthetic scores and attributes as close to the real label on the ground as possible. Bianco, S. et al.~\cite{bianco2018aesthetics} proposed a method to evaluate the aesthetic quality of images with faces by encoding the properties of the whole image and specific aspects of faces. The training test is carried out on AVA-Portraits and the result of LRCC = 0.64 is obtained. Tan, M. et al.~\cite{wettayakorn2018deep} proposed a new fine-tuning sensory model with complete connection and regression layer. The model gives five cores: vivid color, color harmony, lighting, element balance and depth of field. His method combines the retraining of the initial model, the most advanced model for processing images. Their proposed algorithm improves the current level by fine tuning parameters, introducing fully connected layers, and adding regression layers to calculate the numerical score of each focus attribute. The experimental results show that their model helps to reduce the average absolute error (MAE) to 0.211, and benchmark the aesthetic and attribute datasets provided in previous studies. Han et al.~\cite{han2018attribute} proposed a new attribute aware attention model (A3M), which can learn both local attribute representation and global category representation in an end-to-end manner. Hou et al.~\cite{hou2023towards} used the text tags to achieve a higher SRCC on AVA. These excellent works let the models show great ablity on predicting aesthetic scores, but they rely on attribute labels or other tags.

In our work, we take the overall score labels as the only criterion for evaluation and turn different networks for help to extract different information in the picture as the source of attribute features. These mature networks have learned the correct extraction method of attribute features with the help of excellent datasets in their respective fields. We take these features as the attribute input feature of the image aesthetic attribute evaluation network. It is found that the aesthetic attribute score of our model has a high correlation with the corresponding attribute ground-truth in the attribute dataset, which proves the effectiveness of our work.

\begin{figure*}[htbp]
	\includegraphics[width=\linewidth]{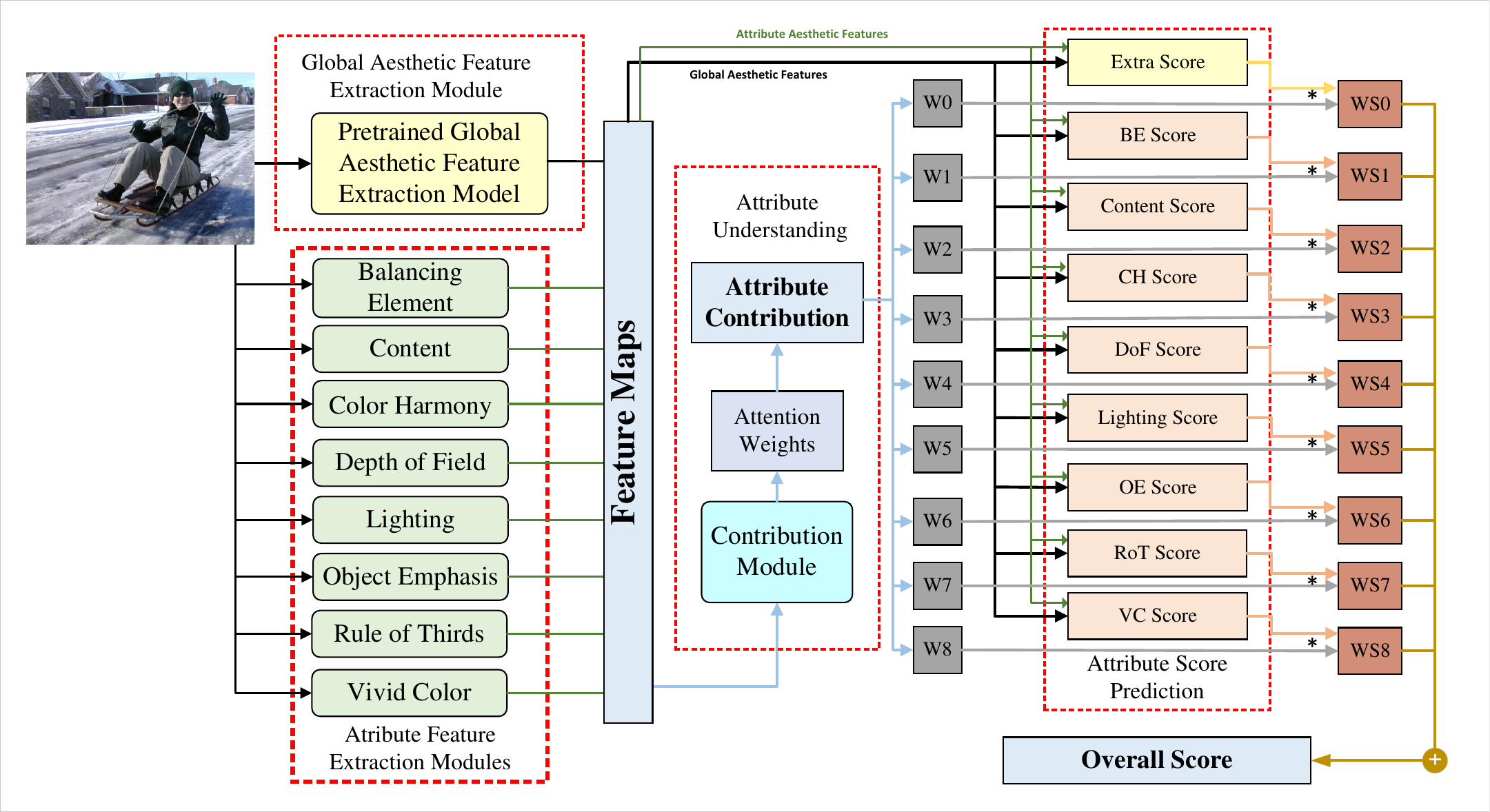}
	\caption{The architecture of our F2S model. It consists of a pre-trained model, attribute feature extraction module, attention module, attribute evaluation module, and attribute contribution module. The pre-trained model is used to extract image aesthetic features. Pre-trained attribute feature extraction networks are adopted for intermediate feature extraction. The attributes we choose are based on AADB dataset labels. Each attribute is integrated with aesthetic features. Each attribute score is further computed from the mixed features and its attribute features. We use the contribution module to compute the contribution of each attribute and adjust the impact of attributes on the overall score. Finally, we can get an overall predicted score and the set of attribute scores.}
	\label{fig:main_model}
\end{figure*}

\section{Proposed Method}
\subsection{Main Architecture}

As shown in Figure~\ref{fig:main_model}, the proposed F2S model consists of five main parts: 1. A main pre-trained network  for global aesthetic feature extraction. 2. A set of attribute feature extraction modules. 3. An attribute understanding contribution module for attribute contribution prediction. 4. The attribute score prediction module.


\subsubsection{Pre-trained Model.}
In our method, we use a main pre-trained network for global aesthetic feature extraction. As shown in Table~\ref{casc_aadb}, our main comparator Leida et al.~\cite{2021lileida} use ResNet50~\cite{he2016deep} as their pre-trained model. So we also use ResNet50 to extract global aesthetic features.

Our commonly used pre-trained models such as ResNet~\cite{he2016deep}, VGG~\cite{simonyan2014very}, and EfficientNet~\cite{tan2019efficientnet} are trained for image classification. Their parameters are trained from image classification datasets such as ImageNet. These models focus on one object in images. When we conduct an image aesthetic evaluation, we need the network to focus on aesthetic features, which may be holistic or texture-based. Therefore, before using the pre-trained model, we use the aesthetic dataset AVA for fine-tuning.

\subsubsection{Attribute Scores.}
Global aesthetic features are stitched with attribute features into a feature map to predict the overall score. In the attribute score module, each attribute has the mixed feature map and its attribute feature as input to predict this attribute score. In our scoring module, we refer to the distribution prediction method of NIMA~\cite{talebi2018nima}, and each attribute scoring module outputs the distribution of the attribute scores. In NIMA~\cite{talebi2018nima}, they use softmax to adjust the output of the last FC layer. 

In the attribute scores, we use global aesthetic features and mixed features to predict the  attribute score. In our view, the aesthetic evaluation of an image depends on many attributes. But in one model, we cannot output all attribute scores. For this reason, we use the ''extra'' attribute to represent other attributes that we didn't consider in one model.

\subsubsection{Attribute Contributions.}

In AADB, the overall score and attribute evaluation are not always independent. The overall score always be high when most attribute evaluations are high. Because of the relationship between overall scores and attribute scores, we propose the concept named attribute contribution. The overall score is determined by attribute scores and attribute contributions. To get the attribute contributions, we use a full-connection layer as the contribution module and take the output as the attribute contributions. The layer output the weight features $y$. We define attribute contributions as
\begin{equation}
	C_{i} = \frac{e^{y_{i}}}{\sum_{i=1}^{N} e^{y_{i}}}
	\label{con:c}
\end{equation}
$y_i$ is the weight features of attribute $i$. $N$ is the number of attributes and a ''extra'' attribute. We use Formula~\ref{con:overall} to explain the relationship among our contribution values, the attribute scores, and the overall score.

\begin{equation}
	S_{overall} = \sum_{i=1}^{N} S_{i} \cdot C_{i}
	\label{con:overall}
\end{equation}

The the contribution module outputs weight vectors. These attribute features are used to predict overall scores rather than attribute scores. So we can use Formula~\ref{con:overall} to train our overall score.

\subsection{Aesthetic Attributes}
\label{ga}

Many aesthetic attributes affect the image's overall aesthetic scores and the aesthetic attributes displayed on different types of images are also different. In architectural photography, the aesthetic attributes include architectural shape and texture. In portrait photography, the aesthetic attributes include posture and expression. These reasons lead to the difficulty that new image aesthetic attribute datasets are proposed.

To prove the value of our work, we design different attribute modules and train our AAEI model in AADB, PCCD, and EVA.

\subsubsection{F2S on AADB}
There are overall scores and 11 aesthetic attribute scores in the AADB dataset. Aesthetic attributes include content, object emphasis, lighting, color harmony, vivid color, depth of field, motion blur, rule of thirds, balancing element, repetition, and symmetry. Among them, motion blur, repetition, and symmetry are mostly 0 in labels. Therefore, we only study 8 aesthetic attributes as Kong et al.~\cite{kong2016photo}, Malu et al.~\cite{2017Learning}, and Leida et al.~\cite{2021lileida} did.

\textbf{Balancing Elements. }
``Balancing elements'' is a very comprehensive feature. In the introduction of AADB, the description of balance elements is "what the image contains balanced elements", which is very brief. In photography, it is the sense of weight produced by human vision on each element in the photo, thus forming a kind of left-right, up-down, or balanced psychology. The sense of weight generated by human vision is through brightness, saturation, eyes, geometric figures, visual guidance, contrast, etc. in the photo elements, which are arranged in the order of visual weight. The brightest part has the heaviest visual weight. The distribution of these elements in the photos is very particular, and people's eyes will quickly judge whether the photos are balanced. It is hard for us to find a way to extract ``Balancing Elements'' features. At last, we use the 3 most related attribute features - ``light'', ``Color'' and ``Object'' as ``Balancing Elements'' features.

\textbf{Content. }
Content includes the scene, things, and even the behavior of the subject in the photo. In AADB, the description of content is "when the image has good/interesting content". This attribute is also an excellent example of the relationship between objective attributes and subjective attributes. The objective content in a photograph includes scenery, architecture, people, animals, etc., and the aesthetic content is interpreted as whether the objective content is good/interesting. In the model, we choose scene segmentation, which is the most detailed analysis of a image, to extract the objective content features. Specifically, we use MaskFormer~\cite{cheng2021maskformer} to extract content features.

\begin{figure}[htbp]
	\centering
	\includegraphics[width=\linewidth]{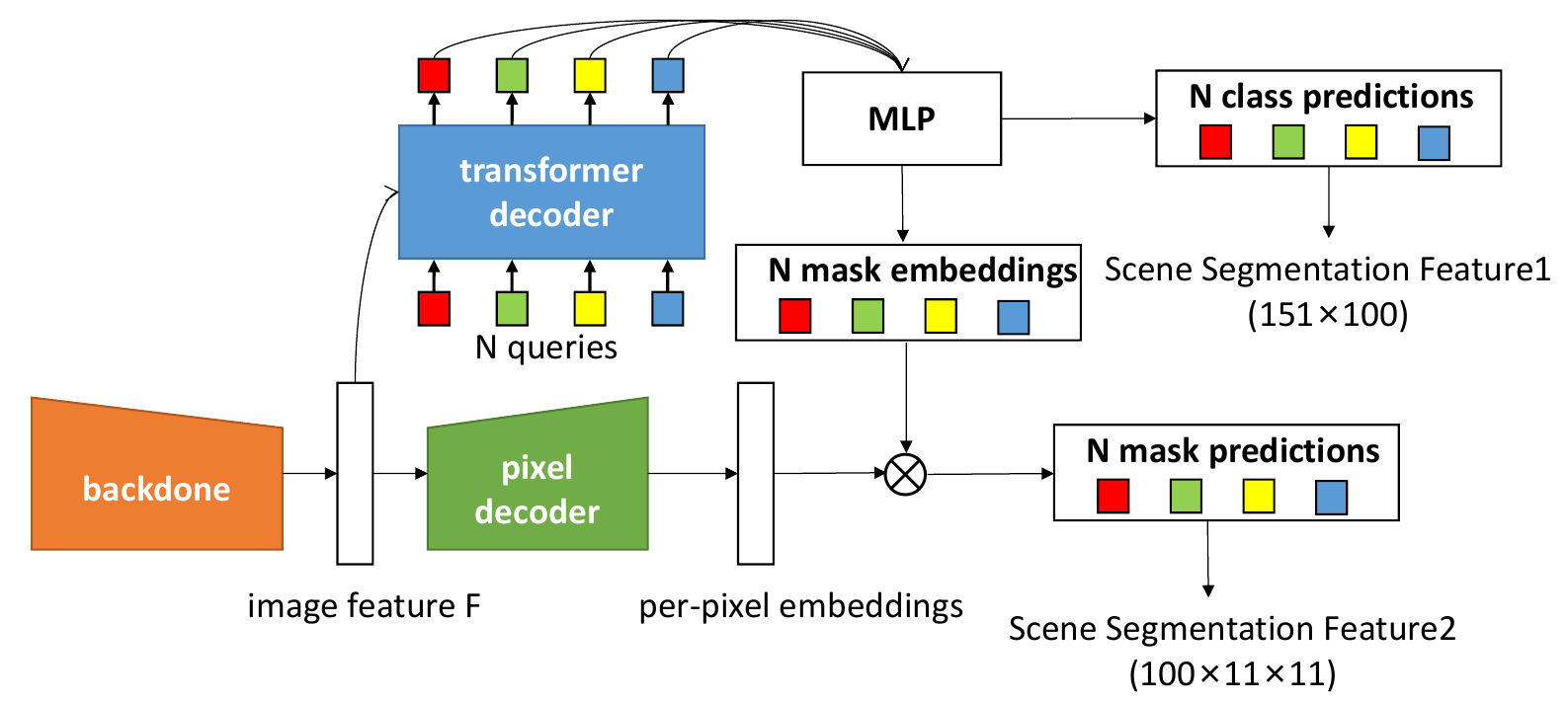}
	\caption{In the MaskFormer model~\cite{cheng2021maskformer}, the scene segmentation features consist of classification prediction features and mask prediction features. To obtain more complete ``Content'' features, the features extracted contain two parts, the ``N class predictions'' part and the ``N mask predictions'' part.}
	\label{fig:ss}
\end{figure}

\textbf{Color Harmony, Vivid Color.}
In AADB, the color attributes are divided into color harmony and vivid color, which are respectively interpreted as "what the overall color of the image is harmonic" and "what the photo has vivid color, not necessarily harmonic color". The two aesthetic features can be said to be different tendencies toward the same image attribute. Color harmony can be understood objectively as whether the color style of the image is uniform, and can be specifically decomposed into the variance of color values in the different fields of a image. As for vivid color. People tend to think that the colors closer to the three-primary colors are more vivid. We count the mean and variance of the hue and saturation in each area of the image. The mean data is used as the features of ``vivid color'' and the variance data is used as the features of ``color harmony''.

\begin{figure}[htbp]
	\centering
	\includegraphics[width=\linewidth]{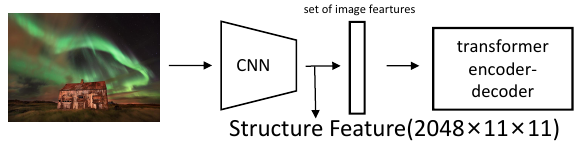}
	\caption{The DETR model~\cite{carion2020end} consists of a CNN feature extraction module and transformer module. Consider that the attribute features will be input into ViT module, the object features come from the CNN module.}
	\label{figdetr}
\end{figure}

\textbf{Object Emphasis. }
In principle, in the model without attribute supervision, the objective attribute features should match the corresponding aesthetic attribute and be mutually exclusive, so that the aesthetic attribute scores are more reasonable. However, the attributes in AADB are not parallel. There are low dimension attributes like ``Light'', ``Color Harmony'' and high dimension attributes like ``Content'', ``Object Emphasis''. At the same time, ``Object Emphasis'' can be included in ``Content''. So we give priority to relevance. DETR~\cite{carion2020end},  a model of object detection with Transformer module. In Figure~\ref{figdetr}, after decoding, we get feature map [$2048\times11\times11$] as attributes to input.

\begin{figure}[htbp]
	\centering
	\includegraphics[width=\linewidth]{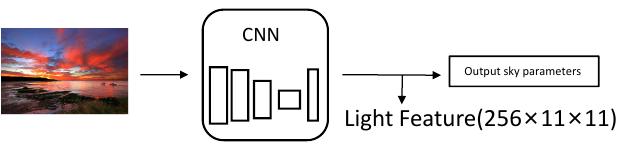}
	\caption{The DOIE model~\cite{hold2017deep} is used to migrate the output sky lighting from an image onto a 3D object. The lighting parameters are used as the ``Light'' features.}
	\label{fig:sky}
\end{figure}

\textbf{Light. }
The ``Light'' attribute extraction model, in Figure~\ref{fig:sky}, is adopted from a technique of outside illumination estimation, proposed by Hold Geoffroy~\cite{hold2017deep}. This model is used to infer illumination parameters those come from [256x11x11] feature map. In order to obtain brightness features more comprehensively, we count the mean and variance of the value in each area of the image.

\begin{figure}[htbp]
	\centering
	\includegraphics[width=\linewidth]{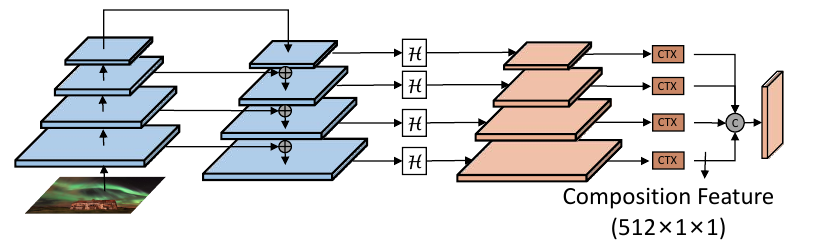}
	\caption{In the DHT model~\cite{zhao2021deep}, the ``CTX'' means the context-aware line detector which contains multiple convolutional layers. The RoT features are the output from the four ``CTX'' modules.}
	\label{fig:dht}
\end{figure}

\textbf{Rule of Thirds (RoT). }
Trisection belongs to the field of composition. In essence, this attribute is an objective attribute. Relevant features can be obtained through composition lines and image segmentation.
Zhao et al.~\cite{zhao2021deep} proposed a one-shot and end-to-end learning framework for line detection. In Figure~\ref{fig:dht}, we get feature map [2048x22x22] as the composition attribute.

\begin{figure}[htbp]
	\centering
	\includegraphics[width=\linewidth]{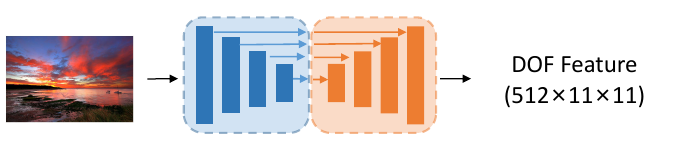}
	\caption{The DoF features come from the Depth network in Monodepth2~\cite{godard2019digging}. }
	\label{fig:depth}
\end{figure}

\textbf{Depth of Field (DOF). }
Godard et al.~\cite{godard2019digging} proposed a method of Self-Supervised Monocular Depth Prediction. The method includes depth network and poses network. In Figure~\ref{fig:depth}, we use Depth network in it to get deep feature [512x11x11].

\subsubsection{F2S on EVA}

We choose the four attributes, visual, composition, quality, and semantic, from EVA as the prediction scores from our F2S. The EVA images come from the AVA dataset. We promise the test data is not in the training data of the global aesthetic feature extraction model. In Table~\ref{mae_eva}, we take the ``object emphasis'' features into one model for comparison.

\textbf{Visual. }
The same as ``Color'' in AADB, we divide images into 16x16 parts and count the mean and variance of the ``H'', ``S'', and ``V'' channels in each part. Than we also use the ``Light'' features used in AADB to describe the ``Visual'' attrbiute.

\textbf{Composition. }
Composition includes the Rule of Thirds (RoT), Triangle, Diagonal, and so on. We use the DHT features~\cite{hold2017deep} that have been used in AADB.

\textbf{Semantic. }
Semantics represents the content of images. We use the MaskFormer features~\cite{cheng2021maskformer} that have bean used in AADB.

\begin{figure}[htbp]
	\centering
	\includegraphics[width=\linewidth]{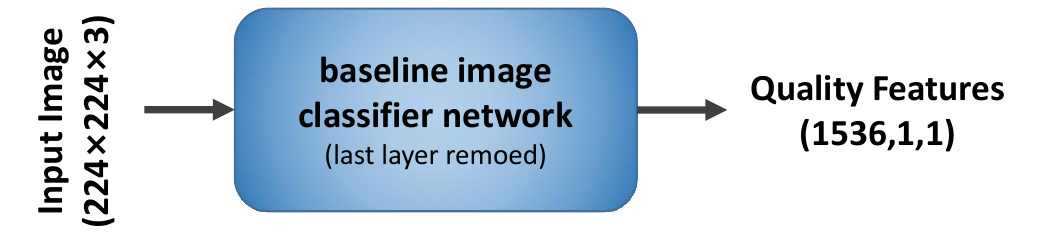}
	\caption{The image quality features come from the network in NIMA~\cite{talebi2018nima}. }
	\label{fig:quality}
\end{figure}

\textbf{Quality. }
Image attribute ''quality`` from EVA is not in AADB. We  NIMA~\cite{talebi2018nima} predict the distribution of human opinion scores using a convolutional neural network. They train the model in TID2013~\cite{ponomarenko2013color}. As shown in Figure~\ref{fig:quality}, we take image quality features from the quality prediction model.

\subsubsection{F2S on PCCD}

PCCD includes ``composition and perspective'', ``color and lighting'', ``subject of photo'', ``depth of field'', ``focus'', and ''use of camera, exposure and speed''. The distinctive attributes are ``focus'' and ``use of camera, exposure and speed''. We calculate the resolution in different image parts as ``focus'' features. We have no ideas to extract features to reflect the ``use of camera, exposure and speed''. The F2S model is free to change different attribute modules, the only requirement is that an attribute module can reflect the aesthetic attribute.

\subsubsection{F2S on portraits}

To show the designable attribute evaluation of F2S, we design a new F2S model to predict attribute scores on portrait images. As shown in Figure~\ref{fig:oe2}, the fourth example is tested on the F2S model trained on portrait images. We use a face recognition network to extract portrait images from the AVA dataset. We get 4000 images for training and 500 images for testing. We promise the test data is not in the training data of the global aesthetic feature extraction model. We choose 4 aesthetic attributes: face lighting, pose, emotion, and appearance. To get the face lighting features, we resort to DPR~\cite{zhou2019deep} which is a face illumination migration network. We use SHN~\cite{newell2016stacked}, a stacked hourglass network for human pose estimation, to get the pose features. As for appearance, the face recognition network is good. We use the face recognition library on OpenCV. At last, we train a light CNN network on fer2013~\cite{goodfellow2013challenges} dataset to extract emotion features. The emotion classifier has over 70\% accuracy.

%
%

\subsection{Loss Function}

We define the loss function as  
\begin{equation}
	L = L_{overall} + \lambda L_{attributes}
\end{equation}
After experimentation, we let $\lambda = 1$. The $L$ has 2 targets, one is to make the overall score close to the ground-truth overall score. So we define the $L_{overall}$ as
\begin{equation}
	L_{overall} = MSELoss(S_{overall},GT_{overall})
	\label{con:loverall}
\end{equation}
We define the $L_{attributes}$ as
\begin{equation}
	L_{attributes} = \sum_{i=1}^{N} MSELoss(S_{i}C_{i} , S_{overall}w_{i})
	\label{con:lattrs}
\end{equation}
The $L_{attributes}$ is defined to let our model output the attribute scores. When we direct use Formula~\ref{con:overall} to train our attribute score, we find that some attribute scores become 0, and other attributes decide the overall score. So we introduce a learnable parameter vector $w$. It is used to guide the training of $S$ and $C$. We believe that the role of a certain attribute in an image is always limited, an attribute cannot completely determine the overall aesthetic value of the image. $w$ makes sure each attribute occupies a relatively fixed position in an image. $w$ is defined as 
\begin{equation}
	w = \frac{e^{S(x)}}{\sum e^{S(x)}} \quad,\quad S(x) = \frac{1}{1+e^{-x}}+\sigma
	\label{con:w}
\end{equation}
$x$ is a learnable parameter vector and $\sigma=1$. $\sigma$ is used to avoid the attributes being invalid.


\section{Experiments}

In the model testing phase, we mainly use SRCC to measure the model capability. Spearman's Rank Correlation Coefficient (SRCC) between the predicted scores and the ground-truth aesthetic scores is an important index. It reflects the correlation. If one image is better than another one, the predicted scores should be higher than another one. We do not use any attribute score labels, so it is hard to predict an accurate score. Our work is more focused on showing comparative relationships.


\subsection{Datasets}

\textbf{AVA~\cite{murray2012ava} and EVA~\cite{kang2020eva}. }AVA dataset is the largest dataset in the field of computational image aesthetics. AVA includes 255500 images from ``DPChallenge.com." Each image will be graded by over 50 people based on a 1-10 rate. EVA dataset has 4070 images with 30 to 40 valid votes each. These images can be fond in the AVA dataset. Each image in EVA has 4 scores, including visual, composition, quality, and semantic.

\textbf{AADB~\cite{kong2016photo}. }In 2016, Kong and others designed a new images' aesthetic datasets, named AADB. AADB showed 10000 images from Flickr website, which was produced by both professional and general photographers. (The ratio of number of professional photographers and general photographers is 1:1.) This datasets also included 11 aesthetic attributes. Because of each image was labeled by only 5 people, the reliability was not as high as AVA.

\textbf{PCCD~\cite{chang2017aesthetic}. } The Photo Critique Captioning Dataset (PCCD) has 4235 images. For each image, professional comments and a rating (from 1 to 10) are given about the following seven aspects: ``general impression'', ``composition and perspective'', ``color and lighting'', ``subject of photo'', ``depth of field'', ``focus'', and ``use of camera, exposure and speed''.

\subsection{Training Process}

The whole trainging process is on Pytorch 1.4 version, i7-7800x and NVIDIA TITAN X.

When we train our global attribute extraction module in the AVA dataset, the training/testing splits of the AVA database can be found at ILGNet~\cite{jin2019ilgnet}. Input images are rescaled to $256\times256$. We use Adam optimizer. Our initial LR is 1e-4 and it will be reduced by 10 times after five times patience in ``ReduceLROnPlateau''. The batch size is 100. We train the model with 40 epochs.

In AVA dataset, ground-truth distribution of human ratings of a given image can be expressed as an empirical probability mass function $p = [p_{s_{1}}, ... , p_{s_{N}}]$ with $s1 \leq s_i \leq s_N$ , where $s_i$ denotes the $i$th score bucket, and $N$ denotes the total number of score buckets. NIMA~\cite{talebi2018nima} proposed this method. In AVA or EVA, $s_1=1$ and $s_N=10$. In AADB, $s_1=0$, $s_N=10$ and $N=11$. As shown in Formula~\ref{con:globalX}, overall scores can be calculated by distribution.
\begin{equation}
	S = \frac{1}{N}\sum_{i=1}^{N} p_{s_i}{s_i} \quad,\quad  \sum p = 1
	\label{con:globalX}
\end{equation}
We use the formula to train our model, its merit is that the scores are limited to between 0 and 1. Whatever we train our model on AADB or EVA or other datasets, the predict module as shown in Figure~\ref{fig:main_model} outputs the distribution of overall scores and attribute scores. As shown in Formula~\ref{con:globalX}, the distribution is converted to a score. Then we calculate MSELoss. If the AVA prediction task, we can output the distribution of overall scores directly and calculate MSELoss with the ground-truth distribution.

As shown in Figure~\ref{fig:main_model}, when we train the F2S model, we prepare the input images and the attribute feature files. Taking AADB as an example, we prepare 8 attribute feature files for each image. We use Adam optimizer. Our initial LR is 1e-4 and it will is multiplied by 0.1 after 5 patience in ``ReduceLROnPlateau''. The batch size is 64. We train the model with 40 epochs.

\begin{figure*}[htbp]
	\includegraphics[width=\linewidth]{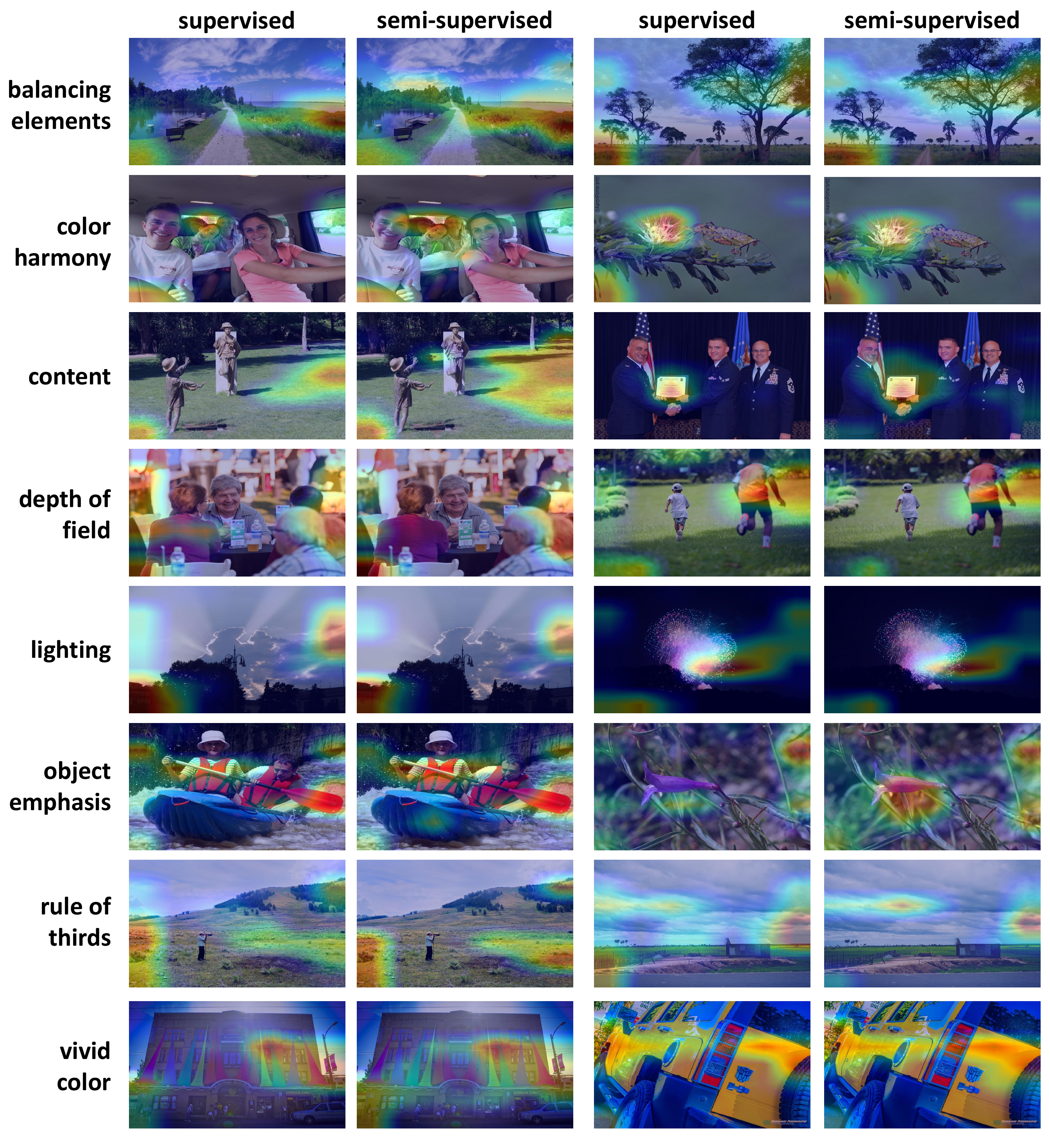}
	\caption{Some examples from AADB. The first and third column images come from our supervised model and the second and fourth column images come from the semisupervised model which has same structure with the supervised model. From the first row to the last row the images have a high score on 'balancing elements',  'color harmony',    'content', 'depth of field', 'lighting',  'object emphasis',  'rule of thirds', 'vivid color'.}
	\label{fig:map}
\end{figure*}

\subsection{Verify Attribute Features}

We use attribute extraction modules to replace the attribute labels. But how can we prove that it is useful? Schultze, S. et al.~\cite{schultze2023explaining} let the features be visualization to show how the model attention the aesthetic attributes. Inspired by them, we designed an experiment to verify the effectiveness. We use two models with the same structure. One is trained with attribute labels, and another is not. When we train the supervised model, we replace $L_{attributes}$ with  
\begin{equation}
	L_{attributes} = \sum^{N}_{i=1}{MSELoss(S_{i},GT_{i})}
	\label{con:su}
\end{equation}
$\lambda=1$.

After training, we use Grad-CAM~\cite{selvaraju2017grad} and compare the CAMs in the last convolution layer between the two models. When we compute CAMs on one attribute, we use single attribute loss of the $L_{attributes}$. As shown in Figure~\ref{fig:map}, the CAMs from the semi-supervised model attention to the same places as from the supervised model. It indicates that the features from related task models can reflect the aesthetic attributes.

\subsection{Performance Analysis}

Some works show good results in SRCC. Most of them use ranking correlation to design their models and loss while our model uses only regression loss. Our models do not use any attribute labels, which means we can address huge costs in the construction of datasets.  As shown in Table~\ref{casc_aadb}, although our model cannot show the best SRCC values, they are not the worst. Our supervised model shows good SRCC values in ``balancing elements'', ``color harmony'', ``content'', ``light'', and ``rule of thirds''. Our semi-supervised model is not the best, but it is also desirable in some attributes. These values are not gaps with those works using attribute labels. The same status happened in the EVA dataset. We reproduced the work from Leida~\cite{2021lileida} and trained their model in the EVA dataset. As shown in Table~\ref{mae_eva}, our work is on par with theirs. When adding more attributes to our model, the SRCC values are higher. In our view, more meaningful attribute features turn into more reasonable attribute scores. We no longer need to take the trouble to score attributes. To test our attribute features, we design an ablation experiment. As shown in Table~\ref{alb1}, our attribute features make our models better. 

\begin{table*}[htb]\small
	\caption{SRCC comparisons on the AADB dataset. Our aesthetic feature extraction model is trained on AVA. The F2S model is trained on the AADB without the supervision of any attribute labels, so we mark  ``\textbf{\XSolid}''. The compared works are trained on attribute labels, so we mark ``\checkmark''. Our method achieves comparable performance. The \textcolor{red}{red} values mean the highest SRCC and the \textcolor{cyan}{cyan} values mean the second highest SRCC.}	\label{casc_aadb}       
	\centering
	\begin{tabular}{p{3cm}p{1.5cm}<{\centering}p{1.5cm}<{\centering}p{1.5cm}<{\centering}p{1.5cm}<{\centering}p{1.5cm}<{\centering}p{1.5cm}<{\centering}}
		\hline\noalign{\smallskip}
		Attributes & Kong et al.~\cite{kong2016photo} & Malu et al.~\cite{2017Learning} & Leida et al.~\cite{2021lileida} &Schultze, Svens et al.~\cite{schultze2023explaining} & Ours (Supervised) & Ours(Semi-supervised) \\
		\noalign{\smallskip}\hline\noalign{\smallskip}
		Labels & \Checkmark & \Checkmark & \Checkmark & \Checkmark & \Checkmark & \textbf{\XSolid}  \\
		\noalign{\smallskip}\hline\noalign{\smallskip}
		Balancing Elements & 0.220  & 0.186 & {0.2514} & 0.236 & \textcolor{red}{0.307} &\textcolor{cyan}{0.286} \\
		Color Harmony & 0.471  & {0.475} & \textcolor{cyan}{0.494} & 0.432 & \textcolor{red}{0.561} & {0.479}  \\
		Content & 0.508 & {0.584}  &\textcolor{red}{0.6117} & 0.494 & \textcolor{cyan}{0.594} &{0.576}\\
		DoF & 0.479 & {0.495} &\textcolor{cyan}{0.5023} & 0.497 & \textcolor{red}{0.530} &0.456 \\
		Light & {0.443} & 0.399 &\textcolor{cyan}{0.467} & 0.401 & \textcolor{red}{0.501} &{0.457} \\
		Object & 0.602 & \textcolor{cyan}{0.666} &\textcolor{red}{0.6768} & 0.627 & 0.661 &0.500 \\
		Rule Of Thirds & {0.225} & 0.178 &0.1731 & \textcolor{cyan}{0.254} & \textcolor{red}{0.283} &{0.237} \\
		Vivid Color & 0.648 & {0.681} &\textcolor{red}{0.7128} & 0.644 & \textcolor{cyan}{0.687} &0.528\\
		\noalign{\smallskip}\hline\noalign{\smallskip}
		Overall Score & 0.678 & 0.689 & \textcolor{red}{0.7317}  & 0.620  & \textcolor{cyan}{0.730} & {0.728}  \\
		\noalign{\smallskip}\hline
	\end{tabular}
\end{table*}

\begin{table*}[htb]\small
	\caption{SRCC comparisons on the EVA dataset. Our aesthetic feature extraction model is trained on AVA. Our F2S models are trained on the EVA without the supervision of any attribute labels. Some of the F2S models use the extra attribute features and output the attribute scores. ``*'' indicates our implementation of Leida et al.~\cite{2021lileida}.}	
	\label{mae_eva}       
	\centering
	\begin{tabular}{p{3cm}p{3cm}<{\centering}p{3cm}<{\centering}p{3cm}<{\centering}p{3cm}<{\centering}}
		\hline\noalign{\smallskip}
		Attributies  & Leida et al.~\cite{2021lileida}* & Ours & Ours(With ``object emphasis'') & Ours(With all AADB attributes)\\ 
		\hline\noalign{\smallskip}
		Visual   & \textcolor{red}{0.72} & 0.587 & 0.604 & 0.648\\
		Composition   & 0.61 & 0.615 & {0.665} & \textcolor{red}{0.679}\\
		Quality  & {0.634} & 0.587 & \textcolor{red}{0.650} & 0.581\\
		Semantic  & {0.565} & 0.467 & {0.629} & \textcolor{red}{0.632}\\
		\noalign{\smallskip}\hline\noalign{\smallskip}
		Overall Score & 0.67 & 0.666 & {0.706} &\textcolor{red}{0.723}\\
		\noalign{\smallskip}\hline
	\end{tabular}
\end{table*}

\begin{table}[htb]\small
	\caption{Ablation study on AADB dataset. The ``Complete'' column is the F2S model with all attribute features. In the ``Ablation'' column, the F2S model replaces the corresponding attribute features in each row with global features. In the ``None'' column, we replaces all attribute features with global features. }
	\label{alb1}       
	\centering
	\begin{tabular}{p{2.5cm} p{1cm}<{\centering} p{1cm}<{\centering}p{1cm}<{\centering}}
		\hline\noalign{\smallskip}
		Attributes & Complete & Ablation & None\\
		\noalign{\smallskip}\hline\noalign{\smallskip}
		BalancingElements &  \textbf{0.286} & 0.276 & 0.256\\
		ColorHarmony & \textbf{0.479} & 0.476 & 0.455\\
		Content & {0.576} & \textbf{0.581} & 0.518\\
		DoF &  \textbf{0.456} & 0.438 & 0.444\\
		Light &  \textbf{0.457} & 0.443 & 0.430\\
		Object &  \textbf{0.500} & 0.455 & 0.460\\
		RuleOfThirds & { 0.237} & 0.214 & \textbf{0.248}\\
		VividColor &  \textbf{0.528} & 0.520 & 0.498\\
		\noalign{\smallskip}\hline
	\end{tabular}
\end{table}

\section{Conclusions}

In this paper, we propose an image aesthetic attribute scoring network that is trained on images with only overall aesthetic labels. The F2S models replace attribute labels with different attribute feature extractors. We analyze the attribute contributions in different attributes. We can design any attributes to the F2S model, it takes some expertise to find useful feature extractors. As shown in Figure~\ref{fig:oe2}, our F2S models, combine with the AVA dataset, can bring many possibilities for the study of the image aesthetic attribute evaluation. The model requires a series of pre-trained attribute feature extractors. Choosing an appropriate feature extractor for a specific aesthetic attribute is essential. Now, our feature extractors are from many other computer vision tasks. The accuracy of those models will affect the performance of the proposed model. In the future, we hope that for various visual domains such as landscapes, statics, and sports, we can discover aesthetic attributes that affect human aesthetic evaluation, improve the aesthetic evaluation ability of the main network, and quantify each aesthetic item in the image.

%
%
%


 
\bibliographystyle{IEEEtran}
\bibliography{IEEEtran}

\vfill

\end{document}